\documentclass[11pt]{article}

\usepackage[final]{acl}

\usepackage{newtx}
\usepackage{latexsym}
\usepackage[T1]{fontenc}
\usepackage[utf8]{inputenc}
\usepackage{microtype}
\usepackage{inconsolata}
\usepackage{graphicx}

\usepackage{amsmath}
\DeclareMathOperator*{\argmax}{arg\,max}
\usepackage{amsthm}
\newtheorem{definition}{Definition}
\usepackage{pifont}
\newcommand{\keypoint}[1]{\vspace{0.2cm}\noindent\textbf{#1}\quad}

\usepackage{wrapfig}
\usepackage{svg}
\usepackage{makecell}
\usepackage{enumitem}
\usepackage{booktabs}
\usepackage{multirow}

\title{The Reward Model Selection Crisis in Personalized Alignment}

\author{
    Fady Rezk$^{1,2}$,
  Yuangang Pan$^2$,
  Chuan-Sheng Foo$^2$,
  Xun Xu$^2$,\\
  \textbf{Nancy F. Chen}$^2$,
  \textbf{Henry Gouk}$^1$,
  \textbf{Timothy Hospedales$^{1,3}$}\\
  $^1$ School of Informatics, University of Edinburgh, UK \\
  $^2$ Agency for Science, Technology and Research (A*STAR), Singapore\\
  $^3$ Samsung AI Research Center, Cambridge, UK
}

\begin{document}
\maketitle

\begin{abstract}
Personalized alignment from preference data has focused primarily on improving personal reward model (RM) accuracy, with the implicit assumption that better preference ranking translates to better personalized behavior. However, in deployment, computational constraints necessitate inference-time adaptation such as reward-guided decoding (RGD) rather than per-user policy fine-tuning. This creates a critical but overlooked requirement: reward models must not only rank preferences accurately but also effectively guide generation. We demonstrate that standard RM accuracy fails catastrophically as a selection criterion for deployment-ready personalized rewards. We introduce \emph{policy accuracy}—a metric quantifying whether RGD-adapted LLMs correctly discriminate between preferred and dispreferred responses—and show that upstream RM accuracy correlates only weakly with downstream policy accuracy (Kendall's $\tau$ = 0.08--0.31). More critically, we introduce \textsc{Pref-LaMP}, the first personalized alignment benchmark with ground-truth user completions, enabling direct behavioural evaluation.
On \textsc{Pref-LaMP}, we expose a complete decoupling between discriminative ranking and generation metrics: methods with 20-point RM accuracy differences produce almost identical output quality, and methods with high ranking accuracy can fail to generate behaviorally aligned responses. These findings reveal that the field has been optimizing for proxy metrics that do not predict deployment performance, and that current personalized alignment methods fail to operationalize preferences into behavioral adaptation under realistic deployment constraints.  In contrast, we find simple in-context learning (ICL) to be highly effective - dominating all reward-guided methods for models $\geq$3B parameters, achieving $\sim$3 point ROUGE-1 gains over the best reward method at 7B scale. 
\end{abstract}

\section{Introduction}

Recent advances in aligning large language models (LLMs) with human preferences have primarily focused on learning from aggregated feedback across diverse user populations \citep{rafailov2023direct,rlhf}. However, preferences are inherently pluralistic—varying across individuals, communities, and contexts \citep{santurkar2023opinionslanguagemodelsreflect,pluralistic_alignment}. This reality motivates \emph{personalized alignment}: adapting model behavior to heterogeneous, sometimes conflicting, user preferences rather than collapsing them into a single consensus objective.

Current personalized alignment research has converged on a common paradigm: collect user-specific preference data (pairwise comparisons), train personalized \textit{ranking}/reward models to capture individual preferences, and assume that better reward models naturally translates to better policies \citep{lore,pal,pref,pdpo,vpl}. The last assumption is likely to break as suggested by Goodhart's law  \cite{elmhamdi2024goodhartslawapplicationvalue}. Unlike standard RLHF, personalized alignment lacks downstream benchmarks that measures policy performance such as MMLU \cite{hendrycks2021measuringmassivemultitasklanguage} and GSM8k \cite{cobbe2021trainingverifierssolvemath}.

\keypoint{Practical Deployment and the End-to-End Perspective}
Per-user policy fine-tuning using personal rewards via RL is computationally infeasible at scale. RL-based personalization requires per-user dynamic adapter management, RL instability mitigation, and orders of magnitude more compute than inference-time alternatives. One key scalable deployment path is \emph{inference-time adaptation} through reward-guided decoding (RGD) \cite{args}, maintaining a single base policy while using personalized rewards to guide generation. Another option is or Best-of-N sampling \cite{bon} but the high latency of BoN makes it unfit for personalization text generation.

This deployment reality demands we adopt an \emph{end-to-end behavioral perspective}: personalized alignment is not merely reward modeling, but the complete process from preference data to actual generation behavior. We propose a key principle:

\begin{quote}
\emph{A personalized alignment method must specify not only how preferences are modeled, but how they are operationalized into behavioral adaptation.}
\end{quote}

A direct corollary of this is that papers proposing reward models are responsible for evaluating if improved RM accuracy translates to improved generation. Current evaluations ignore this responsibility, treating reward modeling and policy adaptation as independent. This obscures whether methods actually achieve their objective: making models generate responses aligned with user preferences.

\keypoint{Our Investigation}
We adopt an end-to-end perspective, studying the complete chain from preference 
modeling to generation behavior. We ask three questions: (1) Does RM accuracy 
predict policy accuracy under RGD? (2) Does policy accuracy predict generation 
quality? (3) How do reward-based alignment methods compare to simpler baselines? To answer these, we introduce (A) policy accuracy, measuring whether RGD scoring assigns higher scores to preferred responses, and (B) \textsc{Pref-LaMP}, a preference learning benchmark with ground-truth user completions enabling direct behavioral evaluation.

\keypoint{Key Findings} Our findings reveal a fundamental selection crisis: practitioners cannot reliably choose deployment-ready methods 
because standard metrics do not predict actual performance.

\textbf{Finding 1:} {Upstream RM accuracy does not predict downstream policy accuracy under RGD (Kendall's $\tau$ = 0.08--0.31). Methods with 20-point RM accuracy differences achieve nearly identical policy performance.}

\textbf{Finding 2:} 
Response \emph{ranking} quality does not predict response \emph{generation} quality. 
On \textsc{Pref-LaMP}, methods with similar generation quality vary dramatically in RM and policy accuracy.

\textbf{Finding 3:} ICL dominates at scale. At 7B parameters, ICL-RAG, with RAG selected preference demonstrations \cite{lamp}, outperforms best personalized reward model by $\sim$3 ROUGE-1 points.

\textbf{Implications:}
For practitioners, use simple ICL-RAG in preference to published personal reward methods. For researchers, take an end-to-end perspective: co-design and co-evaluate reward personalization with policy adaptation strategies and evaluate generation quality as well as ranking. Use \textsc{Pref-LaMP} and develop more benchmarks with ground-truth completions, analogous to GSM8K/MMLU for general RLHF.

\keypoint{Contributions}
To summarize, we (1) contribute \textsc{Pref-LaMP}—the first benchmark with ground-truth user completions, (2) demonstrate the standard RM accuracy metric fails as a selection criterion across three datasets and four scales, (3) demonstrate that a simple ICL baseline outperforms published personal alignment work in end-to-end adaptation, and (4) provide actionable recommendations for practitioners and researchers.

\section{Related Work}

\keypoint{Personalized Alignment}
Recent work has focused on learning user-specific reward models or policies for alignment under limited supervision (e.g., PAL, PReF, LoRE, P-DPO, VPL) \citep{pal,vpl,lore,pref,pdpo}. These approaches largely target reward-modeling accuracy (e.g relative preference ranking) as proxies for personalization quality \cite{pal,lore,pref}. However, such metrics often fail to capture (i) whether RM adaptation translates to downstream policy adaptation, and (ii) whether, under realistic resource-constrained settings, a personalized policy is able to go beyond response ranking and  actually \emph{generate} responses  reflective of a user’s preferences. This evaluation limitation leaves open the null hypothesis that prior  personal alignment results, measured by RM accuracy, are due to unintended overoptimization, also known as reward-hacking \citep{pan2022the}.

\keypoint{Multi-Objective Alignment}
Multi-objective alignment (MOA) addresses the challenge of optimizing language models across multiple known and predefined reward dimensions simultaneously. Unlike personalized alignment, where the goal is to learn individual user preferences under limited supervision, MOA assumes access to distinct reward models for each objective dimension (e.g., helpfulness, harmlessness, factuality) and focuses on finding optimal policy trade-offs among these objectives. Prior work has explored weighted reward optimization \citep{zhou-etal-2024-beyond}, model merging \citep{jang2024personalized,rame2023rewarded}, auxiliary correction models \citep{aligner,metaaligner}, test-time reward-guided decoding \cite{chen2025pad} among other methods \citep{ric}. 

\keypoint{Evaluation Challenges in RLHF}
Existing evaluation practices in RLHF and personalized alignment rely on proxy metrics such as reward-model scores which are susceptible to reward hacking and circularity \citep{tien2023causal,pan2022the}. These methods assess optimization success rather than behavioral quality \cite{wen2025rethinking,gao2022scalinglawsrewardmodel,elmhamdi2024goodhartslawapplicationvalue}. In contrast, our work introduces a framework for \emph{direct behavioral evaluation}, measuring whether generated responses match user-provided completions (Section~\ref{sec:dataset}). Extended discussion of related evaluation pathologies appears in Appendix~\ref{app:extended-related}.
Our work addresses these limitations by introducing direct behavioral evaluation on ground-truth user completions, measuring actual generation quality rather than relying on proxy metrics. 

\keypoint{Inference-Time Alignment}
Reward-guided decoding \cite{args} and Best-of-N sampling \cite{bon} enable policy steering without fine-tuning, making them computationally attractive for personalization. Recent work has explored their effectiveness \cite{wu-2025-comprehensive}, but standard personal alignment evaluation remains limited to reward-based metrics. Our work is the first to systematically evaluate test-time alignment (reward-guided decoding in particular) for personalized alignment with ground-truth behavioral assessment, revealing fundamental limitations in their ability to operationalize user preferences.

\section{Preliminaries and Problem Setup}
\subsection{Personalized Preference Learning}

Consider a preference dataset $\mathcal{D} = \{(u_i, x_i, y_i^{(w)}, y_i^{(l)})\}_{i=1}^N$, where $u_i \in \{1..K\}$ is a user identifier, $x_i$ is a prompt, and $y_i^{(w)}, y_i^{(l)}$ are chosen/winning and rejected/loosing completions. We partition users into $\mathcal{U}_{\mathrm{train}}$ (for learning shared preference structure) and $\mathcal{U}_{\mathrm{adapt}}$ (for evaluating few-shot personalization). Users in $\mathcal{U}_{\mathrm{adapt}}$ are further split into support sets $\mathcal{D}_k^{\mathrm{support}}$ (for adaptation) and query sets $\mathcal{D}_k^{\mathrm{query}}$ (for evaluation). For user $k$, we denote their full dataset as $\mathcal{D}_k = \{(x_i, y_i^{(w)}, y_i^{(l)}) : u_i = k\}$.

\keypoint{Personalized Reward Modeling.} The reward model conditions on user identity: $r_{\theta,z_k}(y \mid x)$, decomposing into shared parameters $\theta$ (general preference structure) and user-specific parameters $z_k$ (individual preferences). Training on $\mathcal{U}_{\mathrm{train}}$ learns both $\theta$ and $\{z_k\}_{k \in \mathcal{U}_{\mathrm{train}}}$. At deployment, for user $k \in \mathcal{U}_{\mathrm{adapt}}$ with support data $\mathcal{D}_k^{\mathrm{support}}$, we adapt:
$$z_k = \mathcal{A}\big(\mathcal{D}_k^{\mathrm{support}}; \theta\big)$$
where $\mathcal{A}$ is the adaptation algorithm.

\subsection{Why Per-User Fine-Tuning Is Infeasible}  \label{sec:cost}

The dominant cost of per-user reinforcement learning is optimization. Each user requires their own rollout generation, reward evaluation, and update loop, a cost that scales linearly in the population and is unchanged by how few parameters are trained: parameter-efficient methods such as LoRA-XS \citep{baazy2024loraxs} reduce per-user state to a few thousand parameters but not the number of RL runs on large models/policies. The few-shot regime that defines personalization is also the regime in which per-user optimization is ill-posed, as tens of preference pairs leave no held-out data on which to tune a KL coefficient or select a checkpoint.

Serving compounds this. Per-user adapters cannot be batched into a single GEMM without gathered kernels and host-to-device paging \citep{slora}. Under reward-guided decoding, introduced next, the policy and reward trunk are usually identical across users and personalization reduces to a product against a low-dimensional user latent, which batches trivially.

\subsection{Deployment via Reward-Guided Decoding}
Given computational constraints prohibiting per-user policy fine-tuning, we deploy personalized alignment through inference-time guidance using Reward-Guided Decoding \cite{args}. 

\keypoint{Reward-Guided Decoding (ARGS).} At each generation step $t$, ARGS scores the top-$k$ tokens $V^{(k)}_t$ retrieved from an off-the-shelf prior LLM policy, $\pi$, using
\begin{equation}\label{eq:args_gen}
\begin{aligned}
\text{score}(v \mid x, y_{<t}; z_k, \lambda)
&= \log \pi(v \mid x, y_{<t}) \\
&\quad + \lambda \cdot r_{\theta, z_k}(v \mid x, y_{<t}),
\end{aligned}
\end{equation}
selecting $y_t = \argmax_{v \in V^{(k)}_t} \text{score}(v)$. This balances base model fluency ($\log \pi$) with personalized alignment ($r_{\theta, z_k}$).

\subsection{The Evaluation Gap}

Standard practice evaluates personalization methods by reward model accuracy, defined below.

\begin{definition}[Reward Model Ranking Accuracy]
\label{def:rm_accuracy}
For user $k$ with evaluation set $\mathcal{D}_k^{\text{query}} = \{(x_i, y_i^{(w)}, y_i^{(l)})\}$, we define the Reward Model Ranking Accuracy as
\begin{equation}\label{eq:acc_rm}
\frac{1}{|\mathcal{D}_k^{\text{eval}}|}
\sum_{\mathcal{D}_k^{\text{eval}}} \mathbb{I}\Big[
r_{\theta, z_k}(y_i^{(w)} \mid x_i)
> r_{\theta, z_k}(y_i^{(l)} \mid x_i)
\Big].
\end{equation}
\end{definition}

This measures pairwise ranking on complete responses of the reward model. Later we will show that this standard metric has several issues and is not predictive of deployment performance.

\section{Policy Accuracy: Measuring Preference Ranking Under RGD}
We introduce a metric quantifying whether the RGD scoring function—not just the reward model in isolation—correctly ranks preferred over dispreferred responses.
\begin{definition}[Policy Ranking Accuracy]
\label{def:policy_accuracy}
Let $s: \mathcal{Y} \times \mathcal{X} \to \mathbb{R}$ be the scoring function used at generation time. The policy accuracy for user $k$ is given by
\begin{equation}\label{eq:acc_policy}
\frac{1}{|\mathcal{D}_k^{\text{query}}|}
\sum_{\mathcal{D}_k^{\text{query}}}\mathbb{I}\Big[
s(y_i^{(w)} \mid x_i)
> s(y_i^{(l)} \mid x_i)
\Big],
\end{equation}
where $y^{(w)}$ and $y^{(l)}$ denote the chosen (winning) and rejected (losing) completions.
\end{definition}
We instantiate $s$ with three scoring functions, each revealing different aspects of the personalization pipeline.

\textbf{Base Policy.} The base policy's length-normalized log-likelihood, its off-the-shelf non-personalized zero-shot ranking ability,
\begin{equation}
s_{\text{base}}(y \mid x) = \frac{1}{|y|}\sum_{t=1}^{|y|} \log \pi(y_t \mid x, y_{<t}).
\end{equation}

\textbf{Global RGD.} RGD with a non-personalized reward model $r_\theta$ trained by aggregating data across all users denoted $s_{\text{global}}(y \mid x)$,
\begin{equation}
\sum_{t=1}^{|y|} \Big[\log \pi(y_t \mid x, y_{<t}) + \lambda \cdot r_{\theta}(y_t \mid x, y_{<t})\Big].
\end{equation}

\textbf{Personalized RGD.} RGD with personalized reward model $r_{\theta,z_k}$ denoted $s_{\text{personal}}(y \mid x; z_k)$,
\begin{equation}
\sum_{t=1}^{|y|}
\Big[
\log \pi(y_t \mid x, y_{<t}) + \lambda \cdot r_{\theta, z_k}(y_t \mid x, y_{<t})
\Big].
\end{equation}

Comparing these reveals: (1) whether reward guidance improves ranking over the base policy, and (2) whether personalization provides gains beyond a global reward model.

\textbf{In-Context Learning.} An alternative personalization mechanism conditions the base policy on user-specific demonstrations $\mathcal{D}_k^{\text{demo}} = \{(x_i, y_i^{(w)},y_i^{(l)}\}\subset\mathcal{D}_k^{\text{support}}$ rather than learning reward parameters. The ICL scoring function, denoted $s_{\text{ICL}}(y \mid x; \mathcal{D}_k^{\text{demo}})$ is
\begin{equation} 
\frac{1}{|y|}\sum_{t=1}^{|y|} \log \pi(y_t \mid \mathcal{D}_k^{\text{demo}}, x, y_{<t}),
\end{equation}
where demonstrations are prepended to the input prompt. Both personalized RGD and ICL leverage user-specific information---$z_k$ versus $\mathcal{D}_k^{\text{demo}}$---but through different mechanisms: learned reward shaping versus direct context conditioning. This allows us to compare whether parametric reward models or demonstration-based adaptation better capture user preferences, particularly as model scale increases.

\textbf{Outstanding Limitation.} Policy accuracy measures how well the scoring function ranks static responses—not whether the policy will actually \emph{generate} outputs that actually align with user preferences. A method might rank existing responses correctly while producing generations that differ substantially from what users would write. This motivates our behavioral evaluation in Section~\ref{sec:dataset}.

\section{Pref-LaMP: A Benchmark for Direct Behavioral Evaluation}\label{sec:dataset}

To enable direct measurement of behavioral alignment without circular reward-based metrics, we introduce \textsc{Pref-LaMP}—a personalized alignment benchmark providing both pairwise preferences and ground-truth user-authored completions.

\keypoint{Dataset Construction}
\textsc{Pref-LaMP} derives from LaMP-5 \cite{lamp}, pairing researchers' abstracts with their titles. Both are author-written, capturing individual style. We construct preferences via hard negative mining: (1) encode abstracts with Qwen3-Embedding-0.6B, (2) retrieve top-$k$ similar abstracts, (3) sample one retrieved abstract as $x$ and use its title as $y^{(l)}$, (4) use original title as $y^{(w)}$. This ensures rejections are topically relevant but different in title formulation\footnote{Human-written rather than LLM-written negatives avoid shortcut learning. Initial LLM-generated rejections let linear probes detect generation artifacts rather than preference signals.}.

\textsc{Pref-LaMP} is the first benchmark enabling direct behavioural evaluation of personalization through user-authored completions, measurable via ROUGE and BERTScore.

\keypoint{Behavioral Alignment Metric} We evaluate end-to-end behavioural alignment by comparing user-generated and personalized model responses.
\begin{definition}[Behavioral Alignment]
\label{def:alignment}
Let $\mathcal{G}: \mathcal{X} \to \mathcal{Y}$ be a generation operator and $\mathcal{S}: \mathcal{Y} \times \mathcal{Y} \to \mathcal{R}$ be a similarity measure. For user $k$ with test set $\mathcal{P}_k = \{(x_j, y_j^{\text{GT}})\}_{j=1}^{M_k}$:
\begin{equation}\label{eq:gteval}
\mathcal{A}_{\mathcal{S}}(\mathcal{G}, k) = \frac{1}{|\mathcal{P}_k|} \sum_{(x_j, y_j^{\text{GT}}) \in \mathcal{P}_k} \mathcal{S}\Big(\mathcal{G}(x_j), y_j^{\text{GT}}\Big)
\end{equation}
\end{definition}

We instantiate $\mathcal{G}$ with ARGS decoding (Eq.~\ref{eq:args_gen}), zero-shot generation and ICL generation. Meanwhile, $\mathcal{S}$ in instantiated with ROUGE-1 (lexical overlap), ROUGE-L (longest common subsequence), and BERTScore-F1 (semantic similarity). This measures whether generated outputs match what users actually write, providing ground-truth assessment of behavioral alignment.

\keypoint{Construction validity.} Pref-LaMP labels are not elicited judgments. The chosen title is the title the author wrote for the given abstract, so the label follows from provenance and requires no annotation; hard-negative mining controls distractor difficulty, not label correctness. To verify that mining produces non-trivial distractors, we compare each rejected title to the chosen title under mining versus a random permutation within the split (Table~\ref{tab:negatives}). Mined negatives are far closer to the ground truth, confirming they are topically matched rather than trivially separable.
\begin{table}[t]
\centering
\small
\begin{tabular}{lccc}
\toprule
Negative & R-1 & R-L & BERTScore \\
\midrule
Mined & 24.6 & 22.2 & 60.0 \\
Random & \phantom{2}4.7 & \phantom{2}4.4 & 43.7 \\
\midrule
$\Delta$ & +19.9 & +17.8 & +16.3 \\
\bottomrule
\end{tabular}
\caption{Similarity of rejected titles to the ground-truth chosen title on the Pref-LaMP evaluation split, under hard-negative mining versus a random permutation of titles within the split. The training split agrees to within $0.01$ on every metric.} \label{tab:negatives}
\vspace{-1em}
\end{table}

\section{Experimental Setup}

\keypoint{Datasets and Models} We consider datasets: 
\textbf{TLDR} \cite{tldr}: Binary stylistic preferences, 10 training users (2,097 prefs/user), 31 adaptation users (100 prefs/user). \textbf{PRISM} \cite{prism}: Pluralistic preferences, 1,232 training users (22.1 prefs/user), 139 adaptation users (14.5 prefs/user). \textbf{Pref-LaMP (ours):} User-authored completions, 485 training users (48.8 prefs/user), 126 adaptation users (49.2 prefs/user).

\begin{table}[t]
    \centering
    \small
    \begin{tabular}{lcc}
        \toprule
        Dataset & Reward Model & Base Policies \\
        \midrule
        TLDR & SmolLM2-180M & 180M, 360M, 1.7B \\
        PRISM & Qwen2.5-0.5B & 0.5B, 1.5B, 3B, 7B \\
        Pref-LaMP & Qwen2.5-0.5B & 0.5B, 1.5B, 3B, 7B \\
        \bottomrule
    \end{tabular}
    \caption{Model configurations.}
    \label{tab:model_configs}
\vspace{-1em}
\end{table}

\keypoint{Models and Methods}
All reward models use LoRA rank 8, trained on $\mathcal{U}_{\text{train}}$ to learn shared $\theta$ and user-specific $\{z_k\}$. We evaluate six personalization methods: LoRE \cite{lore} (learn reward bases and user-specific convex combination), LoRE-Alt (same as LoRE but alternates between bases and user specific parameter gradient steps), PReF \cite{pref} (collaborative filtering), PAL \cite{pal}, VPL \cite{vpl}, MPU/MPU-Avg (a simple baselines of per-user MLPs), and P-DPO \cite{pdpo} (personalized direct preference optimization). Baselines include Global-RM (non-personalized Bradley-Terry using last token embedding as input to RM decoder), Global-RM-V2 (a sequence reward is average reward for all tokens), GenARM \cite{xu2025genarm} (autoregressive RM for token-level guidance), zero-shot generation, ICL (random demonstrations), and ICL-RAG (retrieved demonstrations).

\keypoint{Evaluation Protocol}
We measure: (1) RM accuracy on adaptation users' held-out preferences (Eq.~\ref{eq:acc_rm}), (2) Adaption users' policy accuracy vs prior (no-reward) and global reward baselines (Eq.~\ref{eq:acc_policy}), (3) generation quality on \textsc{Pref-LaMP} via ROUGE-1/L and BERTScore against ground-truth (Eq.~\ref{eq:gteval}), and (4) win rates where each method's RM judges its own outputs versus zero-shot baseline.

\section{Result 1: Reward Model Accuracy Does Not Imply Policy Accuracy}

We first investigate whether reward model accuracy predicts policy accuracy under reward-guided decoding. I.e., whether personal rewards that rank preferences well can guide policies to do the same. 

\begin{table}[t]
    \centering
    \resizebox{1.0\columnwidth}{!}{
    \begin{tabular}{llcccc}
    \toprule
    & \multicolumn{2}{c}{} & \multicolumn{3}{c}{Policy Accuracy} \\
    \cmidrule(lr){4-6}
    &Method & RM Accuracy & 135M & 360M & 1.7B \\
    \midrule
    \multirow{7}{*}{\rotatebox{90}{\textbf{Personalized}}}
    & LoRE & 82.56 $\pm$ 1.12 & 81.09 $\pm$ 0.76 & 81.09 $\pm$ 0.89 & 81.01 $\pm$ 1.10 \\
    &MPU & 76.76 $\pm$ 0.26 & 49.96 $\pm$ 3.24 & 54.36 $\pm$ 7.36 & 55.12 $\pm$ 8.07 \\
    &MPU-Avg & 77.83 $\pm$ 0.19 & 52.91 $\pm$ 4.01 & 52.96 $\pm$ 3.97 & 52.94 $\pm$ 3.98 \\
    &P-DPO & - & - & - & - \\
    &PAL & 87.00 $\pm$ 0.05 & - & - & - \\
    &PReF & 65.27 $\pm$ 8.15 & 46.68 $\pm$ 7.18 & 49.56 $\pm$ 11.77 & 49.65 $\pm$ 11.95 \\
    &VPL & 68.67 $\pm$ 1.94 & 59.78 $\pm$ 4.34 & 60.40 $\pm$ 4.82 & 60.67 $\pm$ 4.92 \\
    \midrule
    \multirow{4}{*}{\rotatebox{90}{\textbf{Baselines}}} &
    GenARM & 53.67 $\pm$ 3.15 & 53.06 $\pm$ 0.02 & 61.56 $\pm$ 0.03 & 62.59 $\pm$ 0.02 \\
    &Global & 73.67 $\pm$ 2.45 & 58.80 $\pm$ 4.79 & 59.58 $\pm$ 1.55 & 60.34 $\pm$ 1.80 \\
    &Global-V2 & 38.66 $\pm$ 1.40 & 53.05 $\pm$ 0.02 & 61.26 $\pm$ 0.07 & 62.54 $\pm$ 0.04 \\
    &Prior & - & 62.00 $\pm$ 0.00 & 61.57 $\pm$ 0.00 &  64.60 $\pm$ 0.00 \\
    \midrule
    &\textit{Pearson $r$} & - & 0.41 & 0.15 & 0.11 \\
    &\textit{Spearman $\rho$} & - & 0.36 & 0.09 & 0.09 \\
    &\textit{Kendall $\tau$} & - & 0.31 & 0.08 & 0.09 \\
    \bottomrule
    \end{tabular}
    }
    \caption{Policy accuracy and reward model performance on TLDR + SmolLM. Base reward model is SmolLM2-135M\label{tab:TLDRsmol}. Top: Personal rewards. Mid: Global alignment.}
\vspace{-1em}
\end{table}

\keypoint{TLDR: Weak Correlation on Simple Data}
We first evaluate the popular TDLR dataset's simple binary style preferences in Table~\ref{tab:TLDRsmol}. The main observation is that \emph{upstream RM accuracy correlates weakly with downstream policy accuracy} (Kendall's $\tau$ = 0.08--0.31), degrading as scale increases (Pearson $r$: 0.41 $\to$ 0.11 from 180M to 1.7B). 

Additionally, most personalization methods (MPU, MPU-Avg, PReF) fail to adapt policies, achieving policy accuracies below the prior baseline (62-65\%), particularly at smaller scales. PAL achieves highest RM accuracy (87.0\%) but doesn't support RGD, while LoRE—with lower RM accuracy (82.6\%)—achieves superior policy accuracy ($\sim$81\%). Overall, LoRE substantially outperforms all competitors as well as prior and global reward policies, demonstrating genuine effectiveness for inference-time adaptation and validating our evaluation framework.

\begin{table*}[t]
\resizebox{1.0\textwidth}{!}{%
    \begin{tabular}{llllllllll}
\toprule
& \multicolumn{7}{c}{\textbf{Personalized}} & \multicolumn{2}{c}{\textbf{Global}} \\
\cmidrule(r){2-8} \cmidrule(l){9-10}

 & LoRE & LoRE-Alt & MPU & MPU-Avg & PAL & PReF & VPL & Global RM & Global RM-v2 \\
\midrule
RM Acc. & \makecell{$ 65.95^{\scriptsize \pm 4.57}$} & \makecell{$ 67.15^{\scriptsize \pm 0.42}$} & \makecell{$ 52.76^{\scriptsize \pm 5.40}$} & \makecell{$ 60.43^{\scriptsize \pm 1.25}$} & \makecell{$ 70.74^{\scriptsize \pm 0.42}$} & \makecell{$ 74.34^{\scriptsize \pm 2.20}$} & \makecell{$68.35^{\scriptsize \pm 2.59}$} & \makecell{$ \textbf{77.94}^{\scriptsize \pm 3.55}$} & \makecell{$ 62.11^{\scriptsize \pm 4.15}$} \\
\bottomrule
\end{tabular}}
\caption{Reward model accuracies of various personal alignment methods  using Qwen2.5-0.5B backbone on PRISM dataset. All personal alignment methods underperform non-personal Global Reward Model V1.}
    \label{tab:prism_main_text_acc}
\end{table*}

\begin{figure*}[t]
    \centering
    \includegraphics[width=2\columnwidth]{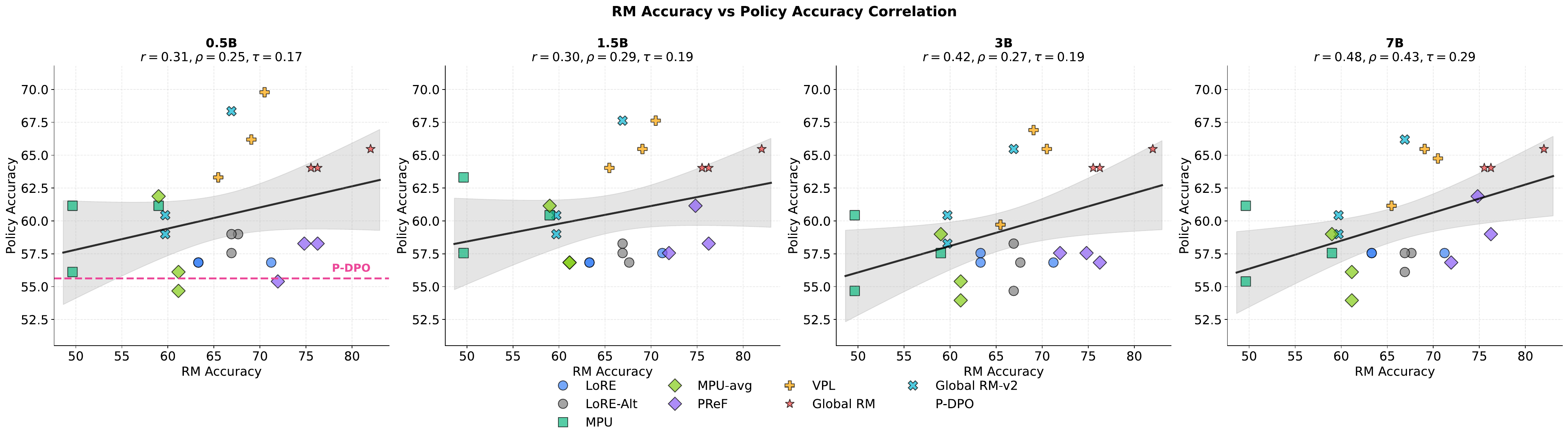}
    \caption{RM vs Policy Accuracy Correlation on PRISM across model scales. Correlations remain consistently weak: Pearson $r$ ranges from 0.30–0.48, Spearman $\rho$ from 0.25–0.43, and Kendall $\tau$ from 0.17–0.29. While correlations slightly strengthen with scale, they remain far below what would be needed for RM accuracy to reliably predict policy performance. Notably, methods with similar RM accuracy can have substantially different policy accuracy and vice-versa, demonstrating that the standard RM metric is not a reliable proxy for deployment performance.}
    \label{fig:rm_policy_corr_prism}
\end{figure*}

\keypoint{PRISM: Personalization Fails on Pluralistic Data}
Results for the more complex PRISM benchmark are summarized in Table \ref{tab:prism_main_text_acc} and Figure \ref{fig:rm_policy_corr_prism}. 
All personalization methods fail in terms of their policy accuracy underperforming the global RM baseline (77.9\% vs. 52.8-74.3\%) in  Table \ref{tab:prism_main_text_acc}. 
Meanwhile, RM-policy correlation, as shown in Fig \ref{fig:rm_policy_corr_prism}, remains weak (Kendall's $\tau$ = 0.17-0.29), slightly strengthening with scale ($r$: 0.31 $\to$ 0.48). Critically, VPL achieves highest policy accuracy (63.8-66.4\%) despite 9.5 points lower RM accuracy than Global RM—a complete ranking inversion. 

In terms of scale, methods show minimal scaling gains, remaining in narrow bands (LoRE-Alt: 57.1-58.5\%, VPL: 63.8-66.4\%).
Unlike TLDR/SmolLM2 where correlations degraded with scale ($r$ = 0.41→0.11), PRISM/Qwen2.5 shows strengthening correlations ($r$ = 0.31→0.48). Whether this reflects dataset differences, model architecture, or their interaction remains unclear. Regardless, even at 7B, correlations remain too weak for choosing methods based on RM accuracy. 

\keypoint{Discussion} Our careful control evaluation shows wide failure of prior personal alignment methods both in terms of beating global alignment baselines, and in terms of the standard metric of RM accuracy not corresponding to downstream policy accuracy. {We attribute this to a mixture of released code not reproducing results, missing non-personal baselines, and inconsistent non-comparable choice of datasets in prior evaluations. See Appendix \ref{app:pref_arch} and \ref{app:lore_arch} for further discussion}.


\textbf{Implication:} Personal RM accuracy does not reflect performance during policy inference and cannot guide choice of reward model for deployment: Methods with 10+ point RM gaps can perform identically as adapted policies; methods with near identical RM accuracy can have 10+ point gaps in policy accuracy; and alignment methods can invert in ranking between reward and policy evaluations. 

\textbf{Recommendation.}
Future personal alignment methods must specify a policy adaptation strategy, and assess downstream policy understanding across multiple datasets and scales—not just upstream personal reward accuracy. The RM-policy disconnect demands new metrics measuring reward models' suitability for guiding generation, rather than pairwise ranking accuracy alone.

\section{Result 2: Preference Discrimination Does Not Imply Generation Quality}

Given the weak correlation between RM accuracy and policy discrimination ability under RGD, we now ask: even when methods achieve high policy accuracy—demonstrably preferring chosen over rejected responses—do they actually generate outputs that behaviorally align with user preferences?

\subsection{Pref-LaMP: Complete Decoupling}
\begin{table*}[t]
\resizebox{1.0\textwidth}{!}{
    \centering
    \small
    \begin{tabular}{llll@{\hspace{1em}}llllllll}
    \toprule
    & \multicolumn{3}{c}{\textbf{Global}} & \multicolumn{8}{c}{\textbf{Personalized}} \\
    \cmidrule(r){2-4} \cmidrule(l){5-12}
    Acc & GenARM & Global RM & Global RM-v2 & LoRE & LoRE-Alt & MPU & MPU-Avg & PAL & PReF & VPL & ICL\\
    \midrule
    RM & \makecell{ $90.04^{\scriptsize \pm 0.69}$ } & \makecell{ $92.24^{\scriptsize \pm 0.25}$ } & \makecell{ $92.26^{\scriptsize \pm 0.23}$ } & \makecell{ $91.97^{\scriptsize \pm 0.10}$ } & \makecell{ $92.11^{\scriptsize \pm 0.18}$ } & \makecell{ $86.15^{\scriptsize \pm 0.39}$ } & \makecell{ $88.02^{\scriptsize \pm 0.40}$ } & \makecell{ $53.77^{\scriptsize \pm 0.28}$ } & \makecell{ $63.79^{\scriptsize \pm 0.88}$ } & \makecell{ $39.65^{\scriptsize \pm 17.76}$ } & \makecell{ NA } \\
    Policy & \makecell{ $76.80^{\scriptsize \pm 8.19}$ } & \makecell{ $66.96^{\scriptsize \pm 0.00}$ } & \makecell{ $84.60^{\scriptsize \pm 1.74}$ } & \makecell{ $67.17^{\scriptsize \pm 0.36}$ } & \makecell{ $67.63^{\scriptsize \pm 0.63}$ } & \makecell{ $67.04^{\scriptsize \pm 0.20}$ } & \makecell{ $64.32^{\scriptsize \pm 4.63}$ } & \makecell{ -- } & \makecell{ $67.13^{\scriptsize \pm 0.29}$ } & \makecell{ $67.19^{\scriptsize \pm 0.40}$ } & \makecell{ $80.20^{\scriptsize \pm 0.69}$ } \\
    \bottomrule
    \end{tabular}
}
\caption{Pref-LaMP preference ranking accuracy for RMs and adapted policies (Qwen2.5-3B). RM personalization does not clearly outperform global RMs for either upstream RM or downstream policy preference ranking.}\label{tab:lamp}
\end{table*}

\begin{figure*}[t]
    \centering
    \includegraphics[width=2\columnwidth]{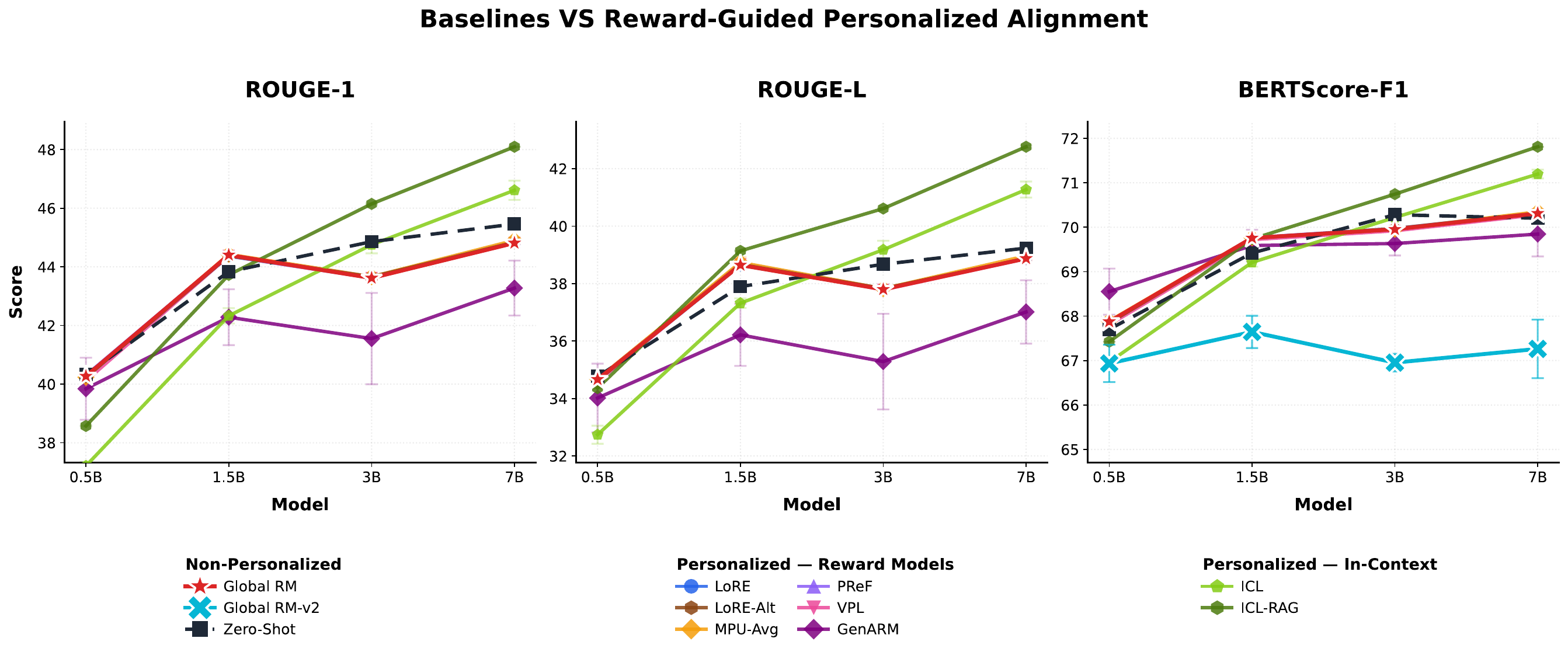}
    \caption{Generation quality (ROUGE-1, ROUGE-L, BERTScore-F1) under RGD across model scales. At 0.5B-1.5B, personalized RMs marginally improve over zero-shot; at 3B-7B, ICL baselines dominate all reward-guided methods.}
    \label{fig:downstream_metrics}
\end{figure*}

We first study Pref-LaMP with preference ranking evaluation in Table~\ref{tab:lamp}. They key observation is that for this challenging task, similarly to PRISM (Table~\ref{tab:prism_main_text_acc}), personal alignment methods struggle to surpass Global RM baselines for both the standard proxy metric of upstream RM accuracy, as well as our downstream policy accuracy.  Only LoRE and LoRE-Alt come close to the global baselines in RM accuracy.

We next move to analysing behavioural generation quality of the policies -- as uniquely enabled by our Pref-LAMP dataset, in Figure~\ref{fig:downstream_metrics} and raw results in Appendix \ref{app:lamp5}. We see that: (1) The top alignment methods for upstream ranking accuracy (Global RM-v2 in Table~\ref{tab:lamp}) tend to underperform in downstream generation quality. (2) A few personal alignment methods can surpass the zero-shot baseline at 1.5B but produce comparative performance to Global RM baseline.
At 3B, with RM accuracies spanning 39.65\% to 92.11\%, all personalized alignment methods ROUGE metrics are concentrated around a 0.05-point spread.
Both these observations reflect decoupling between upstream RM accuracy and downstream behavioural generation. This shows that downstream generation quality evaluation is a crucial missing component of standard evaluation practice. 

Our end-to-end behavioural evaluation also allows direct comparison between existing RM-focused personal alignment approaches and ICL. From Table~\ref{tab:lamp}, we can see that ICL actually achieves better policy preference ranking than the personal RMs. In terms of generation quality, Figure~\ref{fig:downstream_metrics} shows that direct application of ICL, particularly with RAG retrieved demonstrations, surpasses both the baselines and prior personal alignment methods at 3-7B scale, with plain ICL matching zero-shot at 3B and overtaking it at 7B. This suggests that practitioners today should use simple ICL in favour of complex RM-based alignment approaches. 

\begin{table}[t]
\centering
\small
\resizebox{1.0\columnwidth}{!}{
\begin{tabular}{llcccc}
\toprule
&Method & 0.5B & 1.5B & 3B & 7B \\
\midrule
\multirow{3}{*}{\rotatebox{90}{\textbf{Global}}} & GenARM & 100.0\% & 100.0\% & 100.0\% & 99.2\% \\
&Global RM & 92.4\% & 92.1\% & 91.6\% & 92.1\% \\
&Global RM-v2 & 99.0\% & 98.0\% & 92.4\% & 90.8\% \\
\midrule
\multirow{6}{*}{\rotatebox{90}{\textbf{Personalized}}} & LoRE & 92.5\% & 91.7\% & 92.2\% & 92.8\% \\
&LoRE-Alt & 95.0\% & 94.5\% & 94.7\% & 95.5\% \\
&MPU & 84.3\% & 84.2\% & 85.0\% & 85.7\% \\
&MPU-Avg & 80.5\% & 80.0\% & 81.0\% & 82.0\% \\
&PReF & 60.4\% & 59.8\% & 60.9\% & 60.5\% \\
&VPL & 60.5\% & 60.1\% & 59.9\% & 58.6\% \\
\bottomrule
\end{tabular}
}
\caption{Win rate: fraction of examples where each RM judges its own RGD output as better than zero-shot. High win rates reveal severe reward hacking—methods claim near-perfect improvement despite ground-truth metrics showing minimal or negative gains (Fig 2).}
\label{tab:winrate}
\vspace{-2em}
\end{table}

\keypoint{Discussion: Reconciling Prior Claims}
How can we reconcile prior papers' claims of successful RM+RGD based non-personal alignment with the often negative results from our experiments? The evaluation protocol of prior RGD-based analyses involved guiding generation with a RM, and then evaluating the resulting generations using the same RM \cite{args}. The issue with win-rates scored in this way is circularity. If RGD-adaptation can hack the RM (find a 'false positive' response that the RM accepts, while not actually reflecting user preferences), using the same RM to evaluate the result produces overly optimistic results.

To study this protocol, we report win Rate vs zero-shot in Table~\ref{tab:winrate}, which confirms the risk of `circular' evaluation. Using the same RM to guide decoding and judge completions vs a baseline suffers from reward hacking/overoptimization \cite{elmhamdi2024goodhartslawapplicationvalue}.
GenARM claims 100\% improvement over zero-shot despite ROUGE-1 being marginally worse. Global RM claims $\sim$92\% superiority over zero-shot completions while producing outputs identical to methods 20 points lower in RM accuracy. 
The circular evaluation protocol is thus vulnerable to reward hacking, and only appears to solve personal alignment. 

Despite their limitations, ranking held-out preferences (Table~\ref{tab:TLDRsmol},\ref{tab:lamp}) does not suffer from this -- because the RM is not used to generate; neither does our ground-truth evaluation (Figure.~\ref{fig:downstream_metrics}) -- because the RM generation is compared to ground-truth.

\keypoint{Additional Analysis: ICL} We further analyse our strong ICL baselines in terms of number of demonstrations. ICL-RAG improves steadily with demonstrations and scale, reaching $\sim$48 ROUGE-1 at 7B with 8 shots. Larger models show no saturation, effectively leveraging context without reward guidance. This shows that personal alignment is possible and straightforward to implement. This is shown in Figure \ref{fig:scaling_shots_icl}. We provide further ablations in Appendix \ref{app:icl}.
\begin{figure}[t]
    \centering
    \includegraphics[width=\columnwidth]{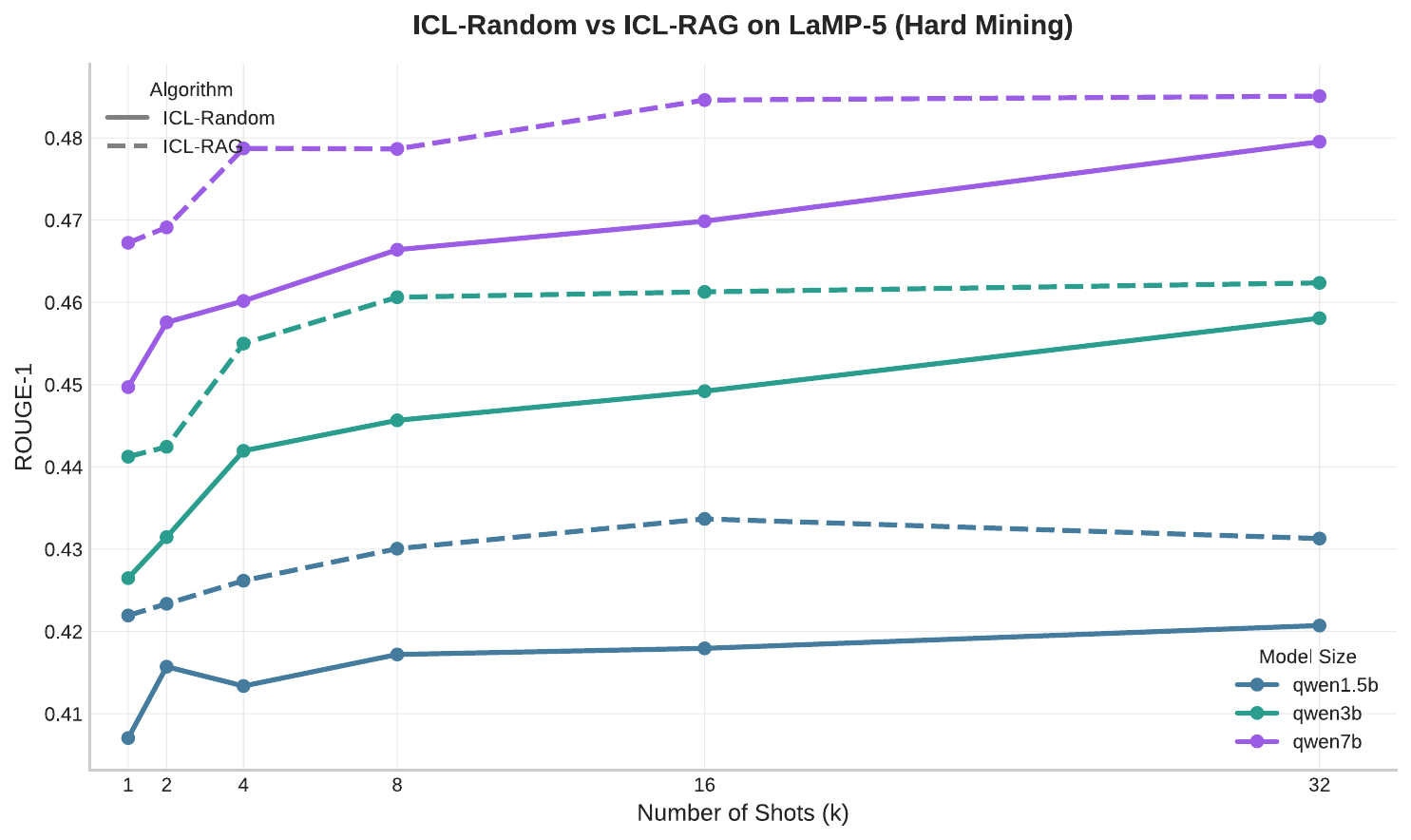}
    \caption{Pref-LAMP generation quality. ROUGE-1 vs. number of in-context examples. ICL-RAG consistently outperforms random ICL, with performance scaling with both model size and context length.}
    \label{fig:scaling_shots_icl}
\end{figure}

\begin{table}[t]
\centering
\small
\begin{tabular}{lccc}
\toprule
\textbf{Scale} & \textbf{RM Acc} & \textbf{Policy Acc} & \textbf{Win Rate} \\
\midrule
\multicolumn{4}{c}{\textit{Kendall's $\tau$ with ROUGE-1}} \\
\midrule
0.5B & $-0.262$ & $-0.266$ & $-0.286$ \\
1.5B & $-0.118$ & $-0.347$ & $-0.269$ \\
3B   & $-0.118$ & $-0.294$ & $-0.149$ \\
7B   & $-0.233$ & $-0.308$ & $-0.262$ \\
\bottomrule
\end{tabular}
\caption{Correlations between metrics and generation quality (ROUGE-1) on Pref-LaMP. All correlations are negligible to negative across model scales, demonstrating no standard metric predicts behavioral alignment.}
\label{tab:metric_correlations}
\vspace{-2em}
\end{table}
\keypoint{No Metric Predicts Generation Quality} RM accuracy, RGD-win rate, and our policy accuracy are all discriminative ranking metrics. Table~\ref{tab:metric_correlations} correlates each against end-to-end generation quality. None is predictive: correlations span $\tau = -0.347$ to $-0.118$ and are negative in all twelve cases, with policy accuracy inverted at every scale. The negative sign is what reward hacking predicts.

\textbf{Takeaway:} No existing metric predicts whether personalization methods will generate aligned outputs. Ground-truth evaluation on user-authored completions is a necessary evaluation component. 

\section{Discussion and Conclusion}

Our findings reveal an evaluation crisis in personal alignment research: RM accuracy is weakly correlated with policy accuracy ($\tau$ = 0.08-0.31), and method rankings can completely invert between upstream and downstream evaluations. Using \textsc{Pref-LaMP}—the first benchmark with ground-truth user completions, we show discriminative metrics fail to predict generation quality (Table~\ref{tab:metric_correlations}): reward models claiming 99\% win rates show no improvement over baselines in ground-truth similarity. The field has been optimizing proxy metrics divorced from deployment objectives.

On a more positive note, we highlight that in contrast to these issues with personal rewards and their evaluations, simple in-context learning dominates reward-guided methods for models $\geq$3B parameters, while being easy and reliable to implement. 

\keypoint{Practical Recommendations.} Practitioners should use ICL with retrieval for 3B+ models; reward modeling adds complexity without benefit at scale. Researchers should: (1) evaluate complete pipelines end-to-end, not just reward model accuracy, (2) include policy accuracy and ground-truth behavioral metrics, (3) test across model scales to detect scale-dependent effects, (4) build behavioral benchmarks with user-authored completions and (5) compare against ICL baselines and focus future research effort on developing such amortized approaches to personal alignment.

\subsection{Limitations}
We focus on RGD because it represents a key scalable deployment path—per-user RL fine-tuning remains computationally infeasible for realistic populations. A fundamental challenge with RGD is that it assumes reward models can be token-wise factorized to provide local guidance at each generation step, which is a known source of error when this assumption is violated \cite{colm}. While GenARM is specifically designed to address this limitation through token-level autoregressive reward training, it still exhibits the same performance gaps we observe across other methods. This suggests the problem runs deeper than factorization alone—the disconnect between preference learning and generation guidance may be fundamental to the inference-time adaptation paradigm.

Our use of three datasets goes beyond most prior work, which often used only one or contrived datasets. However, our results do show some facets of dataset dependence, so future work should aim to establish larger multi-dataset benchmark suits to thoroughly test personalization across more dimensions of interest.

Our behavioral evaluation compares generations to a single ground-truth completion using ROUGE and BERTScore. These metrics are imperfect proxies for preference satisfaction: they penalize valid paraphrases, and on short outputs such as titles they carry high per-example variance. Two properties of our setup limit the impact of this. First, we report a lexical (ROUGE-1/L) and a semantic (BERTScore-F1) metric, and find they order methods similarly, indicating our conclusions are not an artifact of surface-form matching. Second, unlike reward-based win rates, comparison against a held-out author-written completion cannot be gamed by the reward model being evaluated, which is the failure mode we document in Table~\ref{tab:winrate}.

The natural alternative, LLM-as-a-judge conditioned on a user profile, is not yet a sound substitute here. \citet{dong-etal-2024-llm} report around 70\% agreement with human ground truth for a standard personalized judge, and show that verbal uncertainty estimation raises this above 80\% on high-certainty samples, a promising direction we expect to mature. This gain, however, holds only on the subset the judge is confident about, while our conclusions rest on distinguishing methods separated by a few ROUGE-1 points or less. Moreover, a profile-conditioned judge adapts by the same mechanism as ICL, so evaluating ICL with it reintroduces this circularity.

Finally, our conclusions are tied to ARGS and a fixed reward-model backbone across policy scales; whether the decoupling persists under other guidance schemes or larger reward models is open. The methodological claim is independent: personalization deployed through generation should be validated against generation, not ranking accuracy.

\clearpage
\section*{Acknowledgement}
This research/project is supported by the National Research Foundation, Singapore under its National Large Language Models Funding Initiative (AISG Award No: AISG-NMLP-2024-004). Any opinions, findings and conclusions or recommendations expressed in this material are those of the author(s) and do not reflect the views of National Research Foundation, Singapore.
\bibliography{custom.bib}

\clearpage
\appendix
\section{Extended Related Work and Limitations}\label{app:extended-related}
\textbf{Incomparable Accuracy Metrics} 
RLHF and DPO both use pairwise preference accuracy, but these metrics measure fundamentally different things. In RLHF, reward model accuracy measures how well $R_\theta$ ranks response pairs. However, the reward model is not the final artifact---it guides a policy through ARGS or RL fine-tuning. The critical question is: \emph{does the resulting policy generate aligned responses?} Reward model accuracy cannot answer this. A reward model might perfectly rank static pairs while the derived policy fails to generate appropriate responses. In DPO, policy accuracy measures whether $\pi_\phi$ assigns higher probability to preferred responses, but only at the likelihood level---not generation quality. These metrics are not comparable across methods, and neither directly measures the ultimate goal: whether generated outputs align with user preferences.

\textbf{Circular Evaluation Under Frozen Rewards} 
A common RLHF practice adapts policies using reward models, then evaluates by measuring if adapted policies achieve higher rewards than baselines. This creates circularity: the reward model serves as both the training signal and evaluation metric. High performance only confirms the policy learned to exploit the reward model's scoring function---not that it captures actual user preferences. If the reward model is misspecified, this circular evaluation systematically hides the failure. A policy could achieve high reward scores while generating responses users would disprefer, and the evaluation cannot detect this because both training and evaluation use the same potentially-flawed reward model.

\textbf{Proxy-Based Evaluation with LLM-as-a-Judge} 
Recent work uses frontier LLMs as judges, conditioning them on few-shot user examples to rank policy outputs. While appealing, LLM-as-judge remains a learned proxy, not a direct measure of user satisfaction. It provides only relative rankings between methods and cannot quantify whether even the best-ranked method produces satisfactory outputs for individual users.

\textbf{Toward Comprehensive Evaluation} 
These limitations motivate our evaluation framework, which: (1) introduces comparable metrics for both reward model quality and policy understanding, (2) breaks circular evaluation by measuring behavioral alignment against ground-truth user completions rather than reward scores, (3) moves beyond proxies to evaluate actual generation quality, and (4) disentangles where personalization succeeds or fails across the reward modeling, policy guidance, and generation stages.

\section{Raw Results: PRISM}
Please note that we only evaluate on a subset of the test split of PRISM. This is because policy accuracy computation was expensive. Reward model's performance on the full test split is in Table \ref{tab:full_prism_acc}. Global RM still outperforms all other methods so our conclusions in the main paper text does not change. Meanwhile, data plotted in Figure \ref{fig:rm_policy_corr_prism} can be found in Table \ref{tab:prism_policy}.

\begin{table*}[t]
\centering
\tiny
\begin{tabular}{lcccccccc}
\hline
 & LoRE & PReF & VPL & Global RM & Global RM-v2 & MPU & MPU-Avg & LoRE-Alt \\
\hline
RM Accuracy & 65.95 $\pm$ 4.57 & 74.34 $\pm$ 2.20 & 68.35 $\pm$ 2.59 & 77.94 $\pm$ 3.55 & 62.11 $\pm$ 4.15 & 52.76 $\pm$ 5.40 & 60.43 $\pm$ 1.25 & 67.15 $\pm$ 0.42 \\
\hline
\multicolumn{9}{c}{Policy accuracies} \\
\hline
0.5B & 56.83 $\pm$ 0.00 & 57.31 $\pm$ 1.66 & 66.43 $\pm$ 3.24 & 64.51 $\pm$ 0.83 & 62.59 $\pm$ 5.04 & 59.47 $\pm$ 2.91 & 57.55 $\pm$ 3.81 & 58.51 $\pm$ 0.83 \\
\hline
1.5B & 57.07 $\pm$ 0.42 & 58.99 $\pm$ 1.90 & 65.71 $\pm$ 1.81 & 64.51 $\pm$ 0.83 & 62.35 $\pm$ 4.63 & 60.43 $\pm$ 2.88 & 58.27 $\pm$ 2.49 & 57.55 $\pm$ 0.72 \\
\hline
3B & 57.07 $\pm$ 0.42 & 57.31 $\pm$ 0.42 & 64.03 $\pm$ 3.81 & 64.51 $\pm$ 0.83 & 61.39 $\pm$ 3.69 & 57.55 $\pm$ 2.88 & 56.12 $\pm$ 2.59 & 56.59 $\pm$ 1.81 \\
\hline
7B & 57.55 $\pm$ 0.00 & 59.23 $\pm$ 2.53 & 63.79 $\pm$ 2.31 & 64.51 $\pm$ 0.83 & 61.87 $\pm$ 3.81 & 58.03 $\pm$ 2.91 & 56.35 $\pm$ 2.53 & 57.07 $\pm$ 0.83 \\
\hline
\end{tabular}
\caption{Reward model accuracy and policy accuracies per model and method for the PRISM dataset.\label{tab:prism_policy}}
\end{table*}

\begin{table*}
\small
\centering
\begin{tabular}{llllllllll}
\toprule
& LoRE & LoRE-Alt & MPU & MPU-Avg & PAL & PReF & VPL & Global RM & Global RM2 \\
\midrule
RM Accuracy & \makecell{$ 56.53 $ \\ {\scriptsize $\pm 0.15$}} & \makecell{$ 60.33 $ \\ {\scriptsize $\pm 0.57$}} & \makecell{$ 49.66 $ \\ {\scriptsize $\pm 1.13$}} & \makecell{$ 49.07 $ \\ {\scriptsize $\pm 1.46$}} & \makecell{$ 62.51 $ \\ {\scriptsize $\pm 0.30$}} & \makecell{$ 62.73 $ \\ {\scriptsize $\pm 0.64$}} & \makecell{$ 60.31 $ \\ {\scriptsize $\pm 2.98$}} & \makecell{$ 64.51 $ \\ {\scriptsize $\pm 0.41$}} & \makecell{$ 59.18 $ \\ {\scriptsize $\pm 0.74$}} \\
\bottomrule
\end{tabular}%
    \caption{Test set accuracy evaluated across all samples from unseen users. Note: The results reported in the main text use only a single test sample per user due to computational constraints.\label{tab:full_prism_acc}}
    
\end{table*}

\section{Raw Results: \textsc{Pref-LaMP5}}\label{app:lamp5}
Raw results for \textsc{Pref-LaMP5} dataset can be found in Tables \ref{tab:lamp5_Qwen2.5-0.5B-Instruct}, \ref{tab:lamp5_Qwen2.5-1.5B-Instruct}, \ref{tab:lamp5_Qwen2.5-3B-Instruct} and \ref{tab:lamp5_Qwen2.5-7B-Instruct}.
\begin{table*}[t]
\centering
\begin{tabular}{lccccc}
\toprule
Method & RM Acc & Policy Acc & ROUGE-1 & ROUGE-L & BertScore-F1 \\
\midrule
\multicolumn{6}{c}{\textbf{Reward Models + ARGS}} \\
\midrule
GenARM & 90.04 $\pm$ 0.69 & 76.99 $\pm$ 8.05 & 39.84 $\pm$ 1.06 & 34.02 $\pm$ 1.19 & 68.55 $\pm$ 0.52 \\
Global RM & 92.24 $\pm$ 0.25 & 64.95 $\pm$ 0.00 & 40.27 $\pm$ 0.03 & 34.66 $\pm$ 0.02 & 67.87 $\pm$ 0.00 \\
Global RM-v2 & 92.26 $\pm$ 0.23 & 85.44 $\pm$ 1.67 & 30.64 $\pm$ 0.26 & 25.72 $\pm$ 0.06 & 66.93 $\pm$ 0.42 \\
LoRE & 91.97 $\pm$ 0.10 & 65.01 $\pm$ 0.17 & 40.23 $\pm$ 0.03 & 34.63 $\pm$ 0.05 & 67.87 $\pm$ 0.01 \\
LoRE-Alt & 92.11 $\pm$ 0.18 & 65.58 $\pm$ 0.56 & 40.23 $\pm$ 0.00 & 34.63 $\pm$ 0.05 & 67.85 $\pm$ 0.01 \\
MPU & 86.15 $\pm$ 0.39 & 65.16 $\pm$ 0.36 & 40.25 $\pm$ 0.03 & 34.64 $\pm$ 0.03 & 67.86 $\pm$ 0.02 \\
MPU-Avg & 88.02 $\pm$ 0.40 & 62.58 $\pm$ 4.27 & 40.23 $\pm$ 0.05 & 34.68 $\pm$ 0.06 & 67.89 $\pm$ 0.03 \\
PAL & 53.77 $\pm$ 0.28 & nan $\pm$ nan & nan $\pm$ nan & nan $\pm$ nan & nan $\pm$ nan \\
PReF & 63.79 $\pm$ 0.88 & 65.22 $\pm$ 0.53 & 40.30 $\pm$ 0.09 & 34.68 $\pm$ 0.10 & 67.87 $\pm$ 0.02 \\
VPL & 39.65 $\pm$ 17.76 & 65.32 $\pm$ 0.65 & 40.11 $\pm$ 0.32 & 34.59 $\pm$ 0.12 & 67.81 $\pm$ 0.12 \\

\midrule
\multicolumn{6}{c}{\textbf{DPO}} \\
\midrule
\multicolumn{6}{c}{\textbf{Baselines}} \\
\midrule
Zero-Shot & - & - & 40.32 $\pm$ 0.00 & 34.77 $\pm$ 0.00 & 67.70 $\pm$ 0.00 \\
ICL & - & 74.05 $\pm$ 0.48 & 37.21 $\pm$ 0.14 & 32.74 $\pm$ 0.31 & 66.97 $\pm$ 0.13 \\
ICL-RAG & - & - & 38.57 $\pm$ 0.02 & 34.36 $\pm$ 0.01 & 67.42 $\pm$ 0.00 \\
\bottomrule
\end{tabular}
\vspace{0.2cm} \caption{ROUGE scores for Qwen2.5-0.5B-Instruct (mean $\pm$ std across seeds).}\label{tab:lamp5_Qwen2.5-0.5B-Instruct}
\end{table*}

\begin{table*}[t]
\centering
\begin{tabular}{lccccc}
\toprule
Method & RM Acc & Policy Acc & ROUGE-1 & ROUGE-L & BertScore-F1 \\
\midrule
\multicolumn{6}{c}{\textbf{Reward Models + ARGS}} \\
\midrule
GenARM & 90.04 $\pm$ 0.69 & 76.36 $\pm$ 8.58 & 42.28 $\pm$ 0.95 & 36.22 $\pm$ 1.08 & 69.59 $\pm$ 0.36 \\
Global RM & 92.24 $\pm$ 0.25 & 64.84 $\pm$ 0.04 & 44.41 $\pm$ 0.00 & 38.64 $\pm$ 0.02 & 69.76 $\pm$ 0.00 \\
Global RM-v2 & 92.26 $\pm$ 0.23 & 84.90 $\pm$ 1.88 & 31.92 $\pm$ 0.32 & 26.86 $\pm$ 0.11 & 67.64 $\pm$ 0.36 \\
LoRE & 91.97 $\pm$ 0.10 & 64.97 $\pm$ 0.25 & 44.37 $\pm$ 0.01 & 38.64 $\pm$ 0.03 & 69.75 $\pm$ 0.01 \\
LoRE-Alt & 92.11 $\pm$ 0.18 & 65.30 $\pm$ 0.42 & 44.38 $\pm$ 0.02 & 38.66 $\pm$ 0.05 & 69.76 $\pm$ 0.01 \\
MPU & 86.15 $\pm$ 0.39 & 65.03 $\pm$ 0.20 & 44.38 $\pm$ 0.04 & 38.69 $\pm$ 0.03 & 69.77 $\pm$ 0.01 \\
MPU-Avg & 88.02 $\pm$ 0.40 & 62.66 $\pm$ 4.07 & 44.43 $\pm$ 0.12 & 38.74 $\pm$ 0.18 & 69.76 $\pm$ 0.01 \\
PReF & 63.79 $\pm$ 0.88 & 64.90 $\pm$ 0.10 & 44.41 $\pm$ 0.08 & 38.70 $\pm$ 0.11 & 69.76 $\pm$ 0.04 \\
VPL & 39.65 $\pm$ 17.76 & 65.14 $\pm$ 0.55 & 44.37 $\pm$ 0.02 & 38.67 $\pm$ 0.05 & 69.72 $\pm$ 0.07 \\

\midrule
\multicolumn{6}{c}{\textbf{DPO}} \\
\midrule
\multicolumn{6}{c}{\textbf{Baselines}} \\
\midrule
Zero-Shot & - & - & 43.83 $\pm$ 0.00 & 37.89 $\pm$ 0.00 & 69.42 $\pm$ 0.00 \\
ICL & - & 78.35 $\pm$ 0.41 & 42.33 $\pm$ 0.26 & 37.32 $\pm$ 0.17 & 69.21 $\pm$ 0.10 \\
ICL-RAG & - & - & 43.73 $\pm$ 0.04 & 39.14 $\pm$ 0.02 & 69.74 $\pm$ 0.00 \\
\bottomrule
\end{tabular}
\vspace{0.2cm} \caption{ROUGE scores for Qwen2.5-1.5B-Instruct (mean $\pm$ std across seeds).}\label{tab:lamp5_Qwen2.5-1.5B-Instruct}
\end{table*}

\begin{table*}[t]
\centering
\begin{tabular}{lccccc}
\toprule
Method & RM Acc & Policy Acc & ROUGE-1 & ROUGE-L & BertScore-F1 \\
\midrule
\multicolumn{6}{c}{\textbf{Reward Models + ARGS}} \\
\midrule
GenARM & 90.04 $\pm$ 0.69 & 76.80 $\pm$ 8.19 & 41.55 $\pm$ 1.56 & 35.28 $\pm$ 1.66 & 69.63 $\pm$ 0.27 \\
Global RM & 92.24 $\pm$ 0.25 & 66.96 $\pm$ 0.00 & 43.62 $\pm$ 0.01 & 37.80 $\pm$ 0.00 & 69.95 $\pm$ 0.01 \\
Global RM-v2 & 92.26 $\pm$ 0.23 & 84.60 $\pm$ 1.74 & 34.73 $\pm$ 0.63 & 28.58 $\pm$ 0.56 & 66.95 $\pm$ 0.20 \\
LoRE & 91.97 $\pm$ 0.10 & 67.17 $\pm$ 0.36 & 43.63 $\pm$ 0.01 & 37.81 $\pm$ 0.03 & 69.96 $\pm$ 0.00 \\
LoRE-Alt & 92.11 $\pm$ 0.18 & 67.63 $\pm$ 0.63 & 43.66 $\pm$ 0.05 & 37.83 $\pm$ 0.05 & 69.97 $\pm$ 0.03 \\
MPU & 86.15 $\pm$ 0.39 & 67.04 $\pm$ 0.20 & 43.65 $\pm$ 0.05 & 37.82 $\pm$ 0.06 & 69.96 $\pm$ 0.01 \\
MPU-Avg & 88.02 $\pm$ 0.40 & 64.32 $\pm$ 4.63 & 43.66 $\pm$ 0.11 & 37.82 $\pm$ 0.06 & 69.95 $\pm$ 0.01 \\
PReF & 63.79 $\pm$ 0.88 & 67.13 $\pm$ 0.29 & 43.67 $\pm$ 0.09 & 37.83 $\pm$ 0.06 & 69.97 $\pm$ 0.02 \\
VPL & 39.65 $\pm$ 17.76 & 67.19 $\pm$ 0.40 & 43.62 $\pm$ 0.06 & 37.78 $\pm$ 0.03 & 69.92 $\pm$ 0.05 \\

\midrule
\multicolumn{6}{c}{\textbf{DPO}} \\
\midrule
\multicolumn{6}{c}{\textbf{Baselines}} \\
\midrule
Zero-Shot & - & - & 44.86 $\pm$ 0.00 & 38.67 $\pm$ 0.00 & 70.28 $\pm$ 0.00 \\
ICL & - & 80.20 $\pm$ 0.69 & 44.75 $\pm$ 0.29 & 39.18 $\pm$ 0.32 & 70.22 $\pm$ 0.20 \\
ICL-RAG & - & - & 46.15 $\pm$ 0.03 & 40.61 $\pm$ 0.02 & 70.74 $\pm$ 0.00 \\
\bottomrule
\end{tabular}
\vspace{0.2cm} \caption{ROUGE scores for Qwen2.5-3B-Instruct (mean $\pm$ std across seeds).}\label{tab:lamp5_Qwen2.5-3B-Instruct}
\end{table*}

\begin{table*}[t]
\centering
\begin{tabular}{lccccc}
\toprule
Method & RM Acc & Policy Acc & ROUGE-1 & ROUGE-L & BertScore-F1 \\
\midrule
\multicolumn{6}{c}{\textbf{Reward Models + ARGS}} \\
\midrule
GenARM & 90.04 $\pm$ 0.69 & 77.55 $\pm$ 7.57 & 43.28 $\pm$ 0.94 & 37.01 $\pm$ 1.10 & 69.85 $\pm$ 0.50 \\
Global RM & 92.24 $\pm$ 0.25 & 69.54 $\pm$ 0.00 & 44.81 $\pm$ 0.01 & 38.87 $\pm$ 0.02 & 70.31 $\pm$ 0.00 \\
Global RM-v2 & 92.26 $\pm$ 0.23 & 85.08 $\pm$ 1.67 & 35.58 $\pm$ 1.64 & 30.38 $\pm$ 1.29 & 67.26 $\pm$ 0.66 \\
LoRE & 91.97 $\pm$ 0.10 & 69.73 $\pm$ 0.33 & 44.83 $\pm$ 0.02 & 38.89 $\pm$ 0.01 & 70.31 $\pm$ 0.01 \\
LoRE-Alt & 92.11 $\pm$ 0.18 & 69.90 $\pm$ 0.36 & 44.83 $\pm$ 0.04 & 38.88 $\pm$ 0.00 & 70.32 $\pm$ 0.01 \\
MPU & 86.15 $\pm$ 0.39 & 69.69 $\pm$ 0.07 & 44.83 $\pm$ 0.01 & 38.85 $\pm$ 0.04 & 70.32 $\pm$ 0.00 \\
MPU-Avg & 88.02 $\pm$ 0.40 & 66.35 $\pm$ 5.85 & 44.90 $\pm$ 0.11 & 38.96 $\pm$ 0.15 & 70.35 $\pm$ 0.06 \\
PReF & 63.79 $\pm$ 0.88 & 69.56 $\pm$ 0.04 & 44.85 $\pm$ 0.05 & 38.87 $\pm$ 0.02 & 70.32 $\pm$ 0.02 \\
VPL & 39.65 $\pm$ 17.76 & 69.58 $\pm$ 0.04 & 44.81 $\pm$ 0.01 & 38.86 $\pm$ 0.07 & 70.28 $\pm$ 0.05 \\

\midrule
\multicolumn{6}{c}{\textbf{DPO}} \\
\midrule
\multicolumn{6}{c}{\textbf{Baselines}} \\
\midrule
Zero-Shot & - & - & 45.46 $\pm$ 0.00 & 39.23 $\pm$ 0.00 & 70.20 $\pm$ 0.00 \\
ICL & - & 80.51 $\pm$ 0.19 & 46.61 $\pm$ 0.33 & 41.27 $\pm$ 0.28 & 71.20 $\pm$ 0.10 \\
ICL-RAG & - & - & 48.10 $\pm$ 0.01 & 42.76 $\pm$ 0.05 & 71.81 $\pm$ 0.00 \\
\bottomrule
\end{tabular}
\vspace{0.2cm} \caption{ROUGE scores for Qwen2.5-7B-Instruct (mean $\pm$ std across seeds).}\label{tab:lamp5_Qwen2.5-7B-Instruct}
\end{table*}

\section{Architectural and Initialization Enhancements to PReF for Reward-Guided Decoding and Dataset Considerations}\label{app:pref_arch}

\subsection{Motivation for Architectural Modification}

In our work, we leverage the core principles of PReF but introduce a key architectural modification and a novel initialization scheme. These changes are motivated by the need to adapt PReF from a pairwise \textit{preference} model into a pointwise \textit{reward} model, making it suitable for advanced applications such as reward-guided decoding.

The original PReF model is designed to predict a user's preference for one complete response over another. It computes a single score for a \textit{pair} of items, $(r_1, r_2)$. However, reward-guided decoding requires a scalar reward score for a \textit{single, often incomplete, sequence} at each step of the generation process. The original PReF formulation is therefore unsuitable for this task.

To address this, we modified the PReF architecture to explicitly compute a user-specific reward for an individual response, $R(u, r)$. This allows us to score single candidate sequences during decoding.

\subsection{Original vs. Modified Reward Formulation}

The original PReF model calculates a preference score $s$ for a user $u$ and a pair of responses $(r_1, r_2)$ based on the difference of their feature representations:

\[
s(u, r_1, r_2) = \mathbf{u}^\top \big(\phi(r_1) - \phi(r_2)\big)
\]

Here, $\phi$ is a linear head that projects the LLM's response embeddings into a latent feature space, and $\mathbf{u}$ is the user's embedding vector.

Our modified architecture decomposes this calculation into two distinct reward computations:

\[
R(u, r) = \mathbf{u}^\top \phi(r)
\]
\[
s(u, r_1, r_2) = R(u, r_1) - R(u, r_2)
\]

While these two formulations are mathematically equivalent in the final forward pass, this decomposition presents a significant challenge for the model's initialization, which we address with a novel technique.

\subsection{The Initialization Challenge and Our Solution}

The PReF methodology uses Singular Value Decomposition (SVD) on a $(\text{response\_pair} \times \text{user})$ preference matrix to warm-start the model's parameters. A key challenge arises because the SVD process yields a single feature vector, $\mathbf{v}_p$, for each response pair $p$. This vector $\mathbf{v}_p$ serves as a proxy for the latent feature \textit{difference}, $\phi(r_1) - \phi(r_2)$.

The SVD provides no direct information about the individual feature vectors $\phi(r_1)$ and $\phi(r_2)$. To initialize a linear head $\phi$ that operates on individual responses, we leverage the linearity of the projection and work directly in the difference space.

We achieve this with the following direct regression algorithm:

\begin{enumerate}
    \item \textbf{Perform SVD:} Perform SVD on the preference matrix as in the original PReF to obtain the matrix of pairwise feature vectors $U_S$ (representing $\phi(r_1) - \phi(r_2)$ for each pair) and user embeddings $V_S$.
    \item \textbf{Compute LLM Embedding Differences:} For each unique response pair $(r_1, r_2)$ in the training data, compute the difference of their frozen LLM embeddings: $\mathbf{e}_{\text{diff}} = \mathbf{e}(r_1) - \mathbf{e}(r_2)$.
    \item \textbf{Direct Regression on Differences:} Construct a regression problem where the input matrix $X$ contains all LLM embedding differences $\mathbf{e}_{\text{diff}}$, and the target matrix $Y$ contains the corresponding SVD-derived pairwise features $\mathbf{v}_p$ from $U_S$. Solve for the linear head weights $W$ such that:
    \[
    W \cdot (\mathbf{e}(r_1) - \mathbf{e}(r_2)) \approx \mathbf{v}_p
    \]
    \item \textbf{Bias-Free Linear Regression:} A bias-free (intercept-free) regression is used to find the optimal initial weights $W$ for the linear head $\phi$. This approach is mathematically sound because:
    \begin{itemize}
        \item If $\phi(r) = W \cdot \mathbf{e}(r)$, then $\phi(r_1) - \phi(r_2) = W \cdot (\mathbf{e}(r_1) - \mathbf{e}(r_2))$
        \item Learning $W$ from differences is equivalent to learning it from individual features
        \item The lack of bias term means reward values have an arbitrary global offset, which cancels out in pairwise comparisons
    \end{itemize}
\end{enumerate}

\subsection{End-to-End Training}

Following this warm-start initialization of both the user embeddings (from $V_S$) and the linear reward head (from our direct regression method), the model is trained end-to-end using backpropagation. During training, the model computes individual rewards $R(u, r_1)$ and $R(u, r_2)$ for chosen and rejected responses. The Bradley-Terry preference learning loss is then computed:

\[
\mathcal{L} = -\log \sigma(R(u, r_1) - R(u, r_2))
\]

where $\sigma$ is the sigmoid function. Gradients are backpropagated to fine-tune $\phi$, $\mathbf{u}$, and optionally the LLM encoder if not frozen.

This procedure optimizes the true preference learning objective, with the SVD-based initialization serving as a high-quality starting point that accelerates convergence and improves stability. Any imprecision in initialization (such as the arbitrary offset in absolute reward values) is corrected during training. This enhanced methodology preserves the core insights of PReF's SVD-based initialization while adapting its architecture to support reward-guided decoding.

\subsection{PReF's Acknowledged Synthetic Augmentation}

The original PReF implementation uses a synthetically augmented version of PRISM rather than the natural conversational data. The authors state:

\begin{quote}
``However, the original PRISM dataset cannot be used directly because it was collected in a way that prevents overlap between users and prompts, which is necessary for our method. Therefore, we augmented it with synthetic annotations via the protocol described in PERSONA, resulting in 50 user preferences per prompt.''
\end{quote}

They leverage the PERSONA protocol \citep{persona_dataset}, which uses LLMs as judges to simulate user preferences. The code snippet confirms this synthetic generation approach:

\begin{verbatim}
def get_preference_prism(
                    user_description, 
                    prompt, 
                    response_1,
                    response_2):
    prompt1=prompts.PRISM_no_confidence.\
    format(
        prompt=prompt,
        user_description=user_description,
        response_1=response_1,
        response_2=response_2,
    )
    pref1 = get_completion(
        prompt1,
        system_prompt=None,
        model="gpt-4o-mini",
        temp=0.0
        )
\end{verbatim}
Code can be found in github codebase idanshen/PReF\_code on line 78 in file PReF\_code/utils/data.py. This can be found \href{https://github.com/idanshen/PReF\_code/blob/main/PReF\_code/utils/data.py\#L68}{here}.
\subsection{Data Structure Analysis}
Analysis of their dataset reveals a perfectly uniform structure:

Total training samples: 90,450

Unique $(\text{prompt}, r_1, r_2)$ triples: 1,809

Unique users: 1,200

Users per prompt triple: \textbf{Exactly 50} (zero variance)

Average samples per user: 75.4\\
This perfect uniformity demonstrates artificial construction.

\subsection{Real PRISM Conversational Data}

In contrast, the real PRISM dataset \citep{prism} consists of:
\begin{itemize}
    \item Natural multi-turn conversations between users and AI assistants
    \item Real human preferences expressed through dialogue
    \item Sparse preference matrix (unique conversation contexts)
    \item No systematic overlap between users and prompts
\end{itemize}

Our implementation extracts genuine conversational preferences, yielding $\sim$27K training samples.

\subsection{Comparison and Implications}
PReF's ``PRISM'' dataset differs fundamentally from real PRISM data across multiple dimensions. While PReF uses synthetic data generated by GPT-4o-mini with simulated demographic preferences, real PRISM captures authentic human conversations and actual user choices. The synthetic dataset exhibits a dense matrix structure with exactly 50 users per item, enabling high controllability and strong SVD performance, whereas real PRISM data is characterized by sparse, unique contexts with natural variation that yields weaker SVD results. PReF's dataset contains over 90K samples compared to 27K in the real data, but this larger volume comes at the cost of realism—the synthetic patterns may not generalize to authentic human behavior in the way that real PRISM's genuine user interactions do.

\subsection{Implications for Reproducibility and Comparison}

Key points:
\begin{enumerate}
    \item Not an apples-to-apples comparison—PReF’s dense setup differs fundamentally from real sparse data.
    \item SVD initialization performs better in dense synthetic matrices.
    \item Training dynamics differ due to uniform synthetic distribution.
    \item Evaluation on simulated preferences may not generalize to real human data.
\end{enumerate}

\subsection{Methodological Considerations and Design Choices}

PReF’s reliance on dense user-item overlap is intrinsic to collaborative filtering. Sparse real data poses challenges but reflects real-world personalization.

\textbf{Design options:}
\begin{itemize}
    \item \textbf{Match PReF:} Use synthetic PERSONA-style preferences for strong SVD initialization.
    \item \textbf{Use Real Data:} Accept weaker SVD signals, require stronger regularization and robust training.
    \item \textbf{Hybrid:} Augment sparse real data with synthetic overlap.
\end{itemize}

\textbf{Our Choice:} We prioritize authenticity by using real PRISM conversational preferences in their natural sparse form, tackling the more difficult—but more realistic—personalization problem.

\section{LoRe Architecture}\label{app:lore_arch}
LoRE is a pairwise preference learning method introduced in \textit{LoRe: Personalizing LLMs via Low-Rank Reward Modeling}~\citep{lore}. It learns a reward function from preference data, where each datapoint consists of a user input and two responses, one preferred over the other.

Unlike methods that train a binary classifier to predict which response is better, LoRE optimizes a \textbf{logistic loss} over the \textbf{difference of reward values} assigned to the preferred and dispreferred responses.

\subsection{Architecture}

The LoRE architecture consists of two key components:

\keypoint{Feature Extractor} A shared feature extractor $\phi$ (typically a pretrained language model) processes the input $x$ and response $y$ to produce $K$ base reward scores: $R_\phi(x,y) \in \mathbb{R}^K$.

\keypoint{User-Specific Weights} For each user, we learn a low-rank weight vector $w \in \mathbb{R}^K$ that linearly combines these base rewards to produce a personalized scalar reward:
\begin{equation}
R_w(x,y) = w^\top R_\phi(x,y)
\end{equation}

This architecture allows the model to learn a shared representation of reward dimensions through $\phi$, while capturing individual user preferences through the lightweight weight vectors $w$.

\subsection{Original Loss Formulation}

For a preference pair $(x, y^+, y^-)$ where $y^+$ is preferred over $y^-$, the loss uses the difference of personalized rewards:
\begin{equation}
\mathcal{L}_{\text{LoRE}} = \log(1 + \exp(-w^\top[R_\phi(x, y^+) - R_\phi(x, y^-)]))
\end{equation}

This encourages the model to assign a higher personalized reward to the preferred response $y^+$ over the dispreferred one $y^-$.

\subsection{Two Training Algorithms}

The paper introduces two variants:

\keypoint{LoRE} Trains both user-specific weights $w$ and the feature extractor $\phi$ simultaneously in a single optimization step. This approach was used for the TL;DR dataset in the original implementation.

\keypoint{LoRE-Alt} Uses an alternating optimization strategy: for each batch, it takes one gradient step on the user-specific weights $w$ (freezing the feature extractor $\phi$), then one step on the feature extractor $\phi$ (freezing the weights $w$). This approach was used for more complex datasets in the original implementation.

LoRE-Alt also leverages an off-the-shelf reward model (Skywork RM) and includes a regularization term to prevent the learned model from deviating too far from the pretrained baseline. However, since we train our Qwen2.5-0.5B model from scratch without a pretrained reward model, we omit this regularization.

\textit{Note:} The original codebase does not successfully reproduce results on the PRISM dataset.

\subsection{our Variant: Equivalent Log-Sigmoid Loss}

In our implementation, we instead use:
\begin{verbatim}
loss = -F.logsigmoid(reward_diff).mean()
\end{verbatim}
where:
\begin{verbatim}
reward_diff = w.T @ (
    R_phi(x, y^+) - R_phi(x, y^-)
)
\end{verbatim}

This is mathematically equivalent to the original logistic loss, since:
\begin{equation}
- \log(\sigma(x)) = \log(1 + \exp(-x))
\end{equation}

The \texttt{logsigmoid} loss is a numerically stable, PyTorch-friendly implementation of the same core principle. This change does not affect the training dynamics or final optimization target---it is purely an implementation detail.

\subsection{Architectural Equivalence}

In the original LoRE paper, the reward model can be formulated to take the \textbf{difference of features} directly:
\begin{equation}
w^\top[\phi(x, y^+) - \phi(x, y^-)]
\end{equation}

In our implementation, we compute the reward separately on each response using the $K$-dimensional feature extractor, then take the weighted difference:
\begin{equation}
w^\top R_\phi(x, y^+) - w^\top R_\phi(x, y^-)
\end{equation}

These formulations are mathematically identical due to the linearity of the inner product:
\begin{align}
w^\top[R_\phi(x, y^+) - R_\phi(x, y^-)] = \\w^\top R_\phi(x, y^+) - w^\top R_\phi(x, y^-)
\end{align}

This equivalence holds because the personalization layer (the $w$ weights) is linear in the feature space.

\subsection{ARGS Support}

LoRE also supports \textbf{Alignment as Reward-Guided Search (ARGS)}, where generation is guided at decoding time using the learned reward model. In our implementation, we enable ARGS as a runtime decoding strategy by plugging in the learned reward model as a plug-and-play scoring function.

This is implemented by scoring candidate continuations during beam or sampling-based decoding using the personalized reward:
\begin{equation}
R_w(x, y_{\text{candidate}}) = w^\top R_\phi(x, y_{\text{candidate}})
\end{equation}

This allows us to steer generation toward responses that maximize the learned user-aligned reward signal, without requiring reinforcement learning or sampling from a reward-shaped distribution.

\subsection{Known Reproduction Issues}
It is a known issue that LoRe released code does not reproduce their results on the PRISM dataset due to issues in dataset preparation that confalted reported results. \footnote{\url{https://github.com/facebookresearch/LoRe/issues/1}}. This can be found \href{https://github.com/facebookresearch/LoRe/issues/1}{here}. 
\section{Hyperparameters}\label{app:hparams}
\subsection{Reward Modelling}
For the TLDR dataset, all models were trained with a LoRA of rank 8 and LoRA alpha of 16. rsLora was used for initializaiton. The backbone (LoRA module) was $5\times10^{-5}$. Different models decoder used varying hyperparameters as listed below:
\begin{enumerate}
    \item \textbf{LoRe:} decoder LR=0.5, latent dim=2
    \item \textbf{MPU \& MPU-Avg:} decoder LR=$1\times10^{-3}$
    \item \textbf{PAL:} default hyperparameters from original implementation. Topic/Query projectors with LR=$1\times10^{-4}$ and weight decay of 0.01 and user weights with LR=$5\times10^{-3}$ without weight decay. Finally, latent dimension was 2
    \item \textbf{P-DPO:} Number of soft tokens = 8
    \item \textbf{PReF:} decoder LR was $1\times10^{-3}$ and weight decay of 0.02. Latent dimension was 2
    \item \textbf{Vanilla \& Vanilla-V2:} decoder (reward MLP) was trained with LR=$1\times10^{-4}$
    \item \textbf{VPL:} used original implementation hyperparameters. The VAE component was trained with LR=$1\times10^{-4}$. MLP used same LR and weight decay of 0.001
\end{enumerate}
For PRISM, and \textbf{Pref-LaMP5} dataset, same hyperparameters were used with the differences of:
\begin{enumerate}
    \item \textbf{LoRe:} latent dimension = 8
    \item \textbf{LoRe-Alt:} same learning rates but latent dimension = 20
    \item \textbf{PReF:} latent dimension = 8
    \item \textbf{PAL:} latent dimension = 8
\end{enumerate}

\subsection{ARGS}
To find the best weight hyperparameter for ARGS on TLDR and PRISM dataset, we maximized the policy accuracy metric over the seen/train users test split for each trained model. We found that this is necessary since reward models trained using the same algorithm can converge to producing rewards of different magnitudes. For LaMP-5, we used the harmonic mean of rouge-1, inverse perplexity (coherence), and policy accuracy to find the best weight which is subsequently used during generation.

\section{Use of AI}

We used AI assistance in two capacities during this work:

\textbf{Code development:} We used Cursor IDE with AI-assisted code completion during implementation. All AI-generated code suggestions were manually reviewed and verified before integration into the codebase.

\textbf{Writing assistance:} We used large language models to help articulate technical concepts and improve clarity of exposition. The conceptual content, experimental design, results, and conclusions are entirely our own work. AI assistance was limited to rephrasing and refining presentation of ideas we specified.

All scientific claims, experimental results, and intellectual contributions in this paper are the original work of the authors.

\section{Further ICL Ablation}\label{app:icl}
\keypoint{Demonstration format.} The demonstrations in the main text pair each prompt with both the chosen and the rejected response. To isolate the contribution of this contrastive signal, we ablate the demonstration format, replacing each pair with the chosen response alone for the $8$-shots experiments. Results are reported in Table~\ref{tab:demo-format}. Under retrieval the contrast helps consistently: removing the dispreferred response lowers ROUGE-1 from $46.2$ to $45.6$ at 3B and from $48.1$ to $47.3$ at 7B, margins exceeding the across-repetition standard deviation ($\leq 0.03$) by more than an order of magnitude. Under random demonstration selection the effect vanishes: at 3B the chosen-only format is in fact marginally better ($45.1$ vs.\ $44.8$), and the 7B difference ($46.6$ vs.\ $46.5$) lies within run-to-run variance. The contrastive format thus appears to pay off only when the demonstrations are themselves relevant to the query, and its benefit grows with model scale ($+0.5$ at 3B, $+0.8$ at 7B).

\begin{table}[t]
\centering
\resizebox{1.0\columnwidth}{!}{
\begin{tabular}{llcc}
\toprule
Retrieval & Format & 3B & 7B \\
\midrule
\multirow{2}{*}{Random}
  & Preference pairs & 44.75 {\scriptsize$\pm$0.29} & 46.61 {\scriptsize$\pm$0.33} \\
  & Chosen only      & 45.13 {\scriptsize$\pm$0.14} & 46.50 {\scriptsize$\pm$0.14} \\
\midrule
\multirow{2}{*}{RAG}
  & Preference pairs & \textbf{46.15} {\scriptsize$\pm$0.03} & \textbf{48.10} {\scriptsize$\pm$0.01} \\
  & Chosen only      & 45.62 {\scriptsize$\pm$0.02} & 47.26 {\scriptsize$\pm$0.02} \\
\bottomrule
\end{tabular}}
\caption{ROUGE-1 on LaMP-5 for the demonstration-format ablation at $k=8$. \emph{Preference pairs} shows each demonstration with both the preferred and the dispreferred response; \emph{chosen only} shows the preferred response alone. Each entry averages three repetitions; standard deviations are at most $0.003$ for random selection and at most $0.0003$ for retrieval. Best per column in bold.}
\label{tab:demo-format}
\end{table}

\end{document}